\documentclass[conference]{IEEEtran}
\IEEEoverridecommandlockouts

\usepackage{cite}
\usepackage{amsmath,amssymb,amsfonts}
\usepackage{algorithmic}
\usepackage{graphicx}
\usepackage{textcomp}
\usepackage{jabbrv}
\usepackage{xcolor}
\usepackage{multirow}  
\usepackage{booktabs}  

\begin{document}


\title{Towards Unified Neural Decoding of Perceived, Spoken and Imagined Speech from EEG Signals\\

\author{\IEEEauthorblockN{Jung-Sun Lee}
\IEEEauthorblockA{\textit{Dept. of Artificial Intelligence} \\
\textit{Korea University}\\
Seoul, Republic of Korea \\
jungsun\_lee@korea.ac.kr}
\and
\IEEEauthorblockN{Ha-Na Jo}
\IEEEauthorblockA{\textit{Dept. of Artificial Intelligence} \\
\textit{Korea University}\\
Seoul, Republic of Korea  \\
hn\_jo@korea.ac.kr}
\and
\IEEEauthorblockN{Seo-Hyun Lee}
\IEEEauthorblockA{\textit{Dept. of Brain and Cognitive Engineering} \\
\textit{Korea University}\\
Seoul, Republic of Korea  \\
seohyunlee@korea.ac.kr}
}


\thanks{This work was partly supported by Institute of Information \& Communications Technology Planning \& Evaluation (IITP) grant funded by the Korea government (MSIT) (No. RS--2021--II--212068, Artificial Intelligence Innovation Hub, No. RS--2024--00336673, AI Technology for Interactive Communication of Language Impaired Individuals, and No. RS--2019--II190079, Artificial Intelligence Graduate School Program (Korea University)).}
}

\maketitle

\begin{abstract}
Brain signals accompany various information relevant to human actions and mental imagery, making them crucial to interpreting and understanding human intentions. Brain-computer interface technology leverages this brain activity to generate external commands for controlling the environment, offering critical advantages to individuals with paralysis or locked-in syndrome. Within the brain-computer interface domain, brain-to-speech research has gained attention, focusing on the direct synthesis of audible speech from brain signals. Most current studies decode speech from brain activity using invasive techniques and emphasize spoken speech data. However, humans express various speech states, and distinguishing these states through non-invasive approaches remains a significant yet challenging task. This research investigated the effectiveness of deep learning models for non-invasive-based neural signal decoding, with an emphasis on distinguishing between different speech paradigms, including perceived, overt, whispered, and imagined speech, across multiple frequency bands. The model utilizing the spatial conventional neural network module demonstrated superior performance compared to other models, especially in the gamma band. Additionally, imagined speech in the theta frequency band, where deep learning also showed strong effects, exhibited statistically significant differences compared to the other speech paradigms.

\end{abstract}


\section{INTRODUCTION}

Brain-computer interface (BCI) serves as brain-driven communication pathways that convert neural signals into actionable inputs for external systems\cite{kim2015abstract}. In recent years, active BCI has emerged as a next-generation control interface, offering speech-based interaction by directly harnessing the user’s cognitive states and intentions\cite{gifford2022large}. Various types of user input have been studied, including visual imagery, imagined speech\cite{saha2019speak,lee2019towards,si2021imagined,lee2023AAAI}, motor imagery\cite{pei2021tensor,lou2023less}, and motor execution, each presenting unique advantages and limitations. In this paper, We present a novel, integrative BCI paradigm that encompasses perception, imagined speech, whispered speech, and overt speech. This approach holds promise for addressing various limitations in human-computer interaction and provides BCI users with an alternative method of control.
Humans engage in speech production across a variety of real-world contexts. To enable better control of speech in different real-life scenarios, it is essential to collect, analyze, and research data across various speech states. Machine learning techniques for neural decoding have demonstrated significant success, yet they heavily depend on manually engineered features. In contrast, deep learning approaches can directly learn from raw data and execute tasks in an end-to-end manner\cite{kim2023diff}, making them more applicable to real-world scenarios\cite{kim2018discriminative,hang2019cross,han2020classification}. With this in mind, we evaluated the performance of our proposed brain-based input system using a standard deep neural network (DNN) and investigated straightforward yet effective modifications to tailor the networks more closely to our specific paradigm.

DNNs in the BCI domain, such as EEGNet\cite{lawhern2018eegnet}, ShallowConvNet\cite{pan2016shallow}, and filter-bank convolutional network  (FBCNet)\cite{mane2020multi}, typically employ distinct layers of temporal and spatial convolutions. These networks utilize 1D convolutional kernels of fixed sizes to extract temporal, spectral, and spatial features. Temporal kernels are often chosen heuristically, while spatial kernels are applied uniformly across all channels. This architecture has been highly effective in conventional BCI paradigms, such as visual imagery, motor imagery, and motor execution, which predominantly involve sensory or motor-related signals. Nevertheless, the complex nature of EEG signals demands multi-scale kernels to interpret information across various temporal scales\cite{lee2020continuous,salami2022eeg}. Additionally, due to the effects of volume conduction, EEG signals exhibit redundancy between electrodes, resulting in low spatial resolution. Moreover, current feature extraction methods, which are based in Euclidean space, cannot accurately capture the complex relationships between multiple electrodes, necessitating additional spatial-based features like connectivity methods, phase locking value (PLV), phase lag index (PLI), and coherence, which contain topological spatial information of the brain\cite{jeong2019classification}.

\section{MATERIALS AND METHODS}
\subsection{Dataset Description}

\begin{figure*}[t]
    \centering
    \includegraphics[width=1\textwidth]{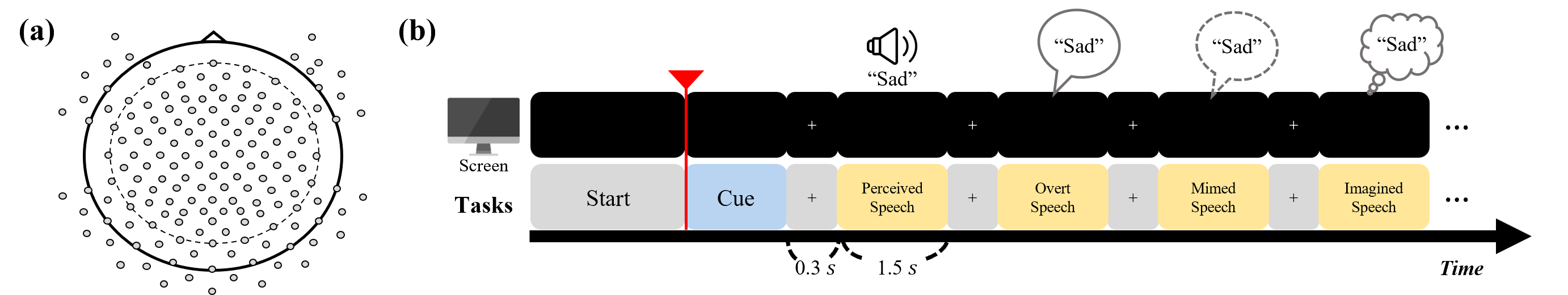} 
    \caption{(a) Placement of 128 electrodes in the 10-5 international standard system. (b) Experimental paradigm for recording EEG signals during four speech states in words. Following the cue, a 1.5-second interval is allocated for perceived speech, during which the participant listens to an auditory stimulus before transitioning to subsequent tasks.}
    \label{fig:paradigm}
\end{figure*}

The study involved ten healthy participants, six males and four females, with a mean age of 23.7 years (SD = 3.23). None of the participants had a history of claustrophobia or hearing impairment, ensuring that the experimental conditions did not induce discomfort or auditory bias. Brain signals were acquired using a 128-channel EEG cap with active Ag/AgCl electrodes that followed the international 10-5 system for electrode placement. The FCz and FPz channels were designated as the reference and ground electrodes. EEG signals were captured using Brain Vision/Recorder software (Brain Products GmbH, Germany) and processed using MATLAB 2023a. A sampling rate of 1,000 Hz was employed to capture EEG recordings, facilitating the detailed examination of dynamic processes inherent in different speech paradigms. 

\subsection{Experimental Setup}

The experimental paradigm investigated neural correlates across different speech conditions, including perceived, overt, whispered, and imagined speech. The dataset was organized into 20 distinct word classes, divided into five categories, each containing four words. The words were selected to cover a range of emotional, natural, and abstract concepts. The speech conditions included perception, overt speech, whispered speech, and imagined speech. 

The word categories comprised emotions, natural objects, animals, artificial objects, and abstract nouns. Examples of words in the emotion category are `Sad,' `Amused,' `Positive,' and `Disappointed.' For natural objects, the words included `Peach,' `Mango,' `Strawberry,' and `Watermelon.' The animal categories included `Horse,' `Tiger,' `Buffalo,' and `Alligator.' Artificial objects were represented by `House,' `Notebook,' `Apartment,' and `Television.' Lastly, the abstract nouns included `Death,' `Weather,' `January,' and `Conversation.'

The data collection was conducted over three days. On the first day, 20 trials per word class were recorded, totaling 400 trials. On the second day, another 20 trials per word class were recorded, resulting in an additional 400 trials. On the third day, 60 trials per word class were recorded, totaling 1,200 trials. Altogether, 2,000 trials were recorded, providing a robust dataset for analyzing the neural dynamics associated with different speech conditions and word categories. 

\subsection{Signal Preprocessing}

EEG signals were analyzed using MNE-Python and MATLAB with BBCI toolbox. The EEG data were subjected to a series of pre-processing steps to ensure high-quality signals for analysis. First, a band-pass filter was applied to retain frequencies between 0.5 and 125 Hz. A notch filter was also implemented at 60 Hz and 120 Hz to remove power line noise and its harmonics. Channels with poor signal quality, such as those affected by electrooculography artifacts, were identified and rejected. The data were then segmented into 1.5-second epochs without overlap, resulting in a total of 1,200 trials. These trials were evenly distributed across 20 classes, with 60 trials per class.

Following the short-time Fourier transform (STFT) process, we extracted band power features by summing the power within specific frequency bands: the delta (\(\delta\), 1--4 Hz), theta (\(\theta\), 4--8 Hz), alpha (\(\alpha\), 8--12 Hz), beta (\(\beta\), 12--30 Hz), and gamma (\(\gamma\), 30--45 Hz) bands. This procedure generated a feature matrix of dimensions (number of time windows $\times$ 5 bands) for each EEG channel. Additionally, the data from each channel were converted into power spectral density using the STFT with a sliding window of one second. These five band power features, alongside the connectivity metrics discussed later, were utilized as input for subsequent analyses to decode neural patterns associated with different cognitive states and speech paradigms.

\subsection{Connectivity Decoding Model}

\begin{table*}[t]
    \centering
    \caption{Comparison of the accuracy performance of conventional models based on connectivity methods across frequency bands.} 
    \renewcommand{\arraystretch}{1.5} 

    \begin{tabular}{c c c c c c c c}
        \toprule
        \textbf{Feature} & \textbf{Model} & \multicolumn{5}{c}{\textbf{Frequency Bands (Hz)}} \\
        \cmidrule(lr){3-8}
         & & \textbf{Delta} & \textbf{Theta} & \textbf{Alpha} & \textbf{Beta} & \textbf{Gamma} & \textbf{Total}\\
        \midrule
        \multirow{3}{*}{PLV} & EEGNet\cite{lawhern2018eegnet} & 42.96 $\pm$ 2.11 & 46.87 $\pm$ 2.39 & 41.39 $\pm$ 1.26 & 48.42 $\pm$ 1.15 & \textbf{48.52 $\pm$ 1.69} & \textbf{53.20 $\pm$ 1.84} \\
         & ShallowConvNet\cite{pan2016shallow} & 33.81 $\pm$ 1.29 & \textbf{38.05 $\pm$ 2.21} & 32.21 $\pm$ 2.55 & 34.74 $\pm$ 2.72 & 36.05 $\pm$ 1.83 & \textbf{43.51 $\pm$ 1.75} \\
         & FBCNet\cite{mane2020multi} & 43.31 $\pm$ 1.31 & 48.93 $\pm$ 1.05 & 44.81 $\pm$ 2.36 & 47.8 $\pm$ 1.81 & \textbf{49.05 $\pm$ 1.45} & \textbf{54.91 $\pm$ 1.37} \\
        \midrule
        \multirow{3}{*}{PLI} & EEGNet\cite{lawhern2018eegnet} & 44.85 $\pm$ 1.25 & 49.12 $\pm$ 1.79 & 44.25 $\pm$ 1.74 & 48.42 $\pm$ 1.15 & \textbf{50.84 $\pm$ 1.97} & \textbf{56.94 $\pm$ 1.60} \\
         & ShallowConvNet\cite{pan2016shallow} & 35.49 $\pm$ 1.82 & 38.60 $\pm$ 2.33 & 36.29 $\pm$ 1.11 & 37.6 $\pm$ 2.15 & \textbf{39.12 $\pm$ 1.71} & \textbf{45.10 $\pm$ 2.24} \\
         & FBCNet\cite{mane2020multi} & 45.36 $\pm$ 1.65 & 51.11 $\pm$ 1.29 & 46.85 $\pm$ 2.43 & 49.82 $\pm$ 2.04 & \textbf{51.43 $\pm$ 1.37} & \textbf{54.51 $\pm$ 1.86} \\
        \bottomrule
    \end{tabular}
    \label{tab:freq_band_comparison}
\end{table*}

To quantify the functional connectivity across various frequency bands during perceived, overt, whispered, and imagined speech paradigms, we employed two metrics: PLV\cite{wang2019phase} and PLI\cite{huang2023spatio}. By constructing networks using these two metrics, we investigated the different connection patterns among the four speech patterns.

PLV was utilized to measure the average phase difference between pairs of EEG time series. The instantaneous phases of the signals were obtained using the Hilbert transform. For signals \( x_n \) and \( x_t \) at time point \( k \), the PLV is defined as:

\[
PLV_{n,t} = \left| \frac{1}{M} \sum_{k=0}^{M-1} e^{i(\phi_n(k) - \phi_t(k))} \right|,
\]

\noindent where \( \phi_n(k) \) and \( \phi_t(k) \) represent the instantaneous phases of signals \( x_n \) and \( x_t \) at time point \( k \), respectively, and \( M \) is the total number of samples. The PLV ranges from 0 to 1, with values closer to 1 indicating strong phase synchronization between the signals. This measure provides insight into the temporal coherence and functional connectivity between different EEG channels.

PLI was employed to quantify the average phase lead or lag between two EEG time series by analyzing the sign function of phase differences. Specifically, the PLI between signals \( x_n \) and \( x_t \) at time point \( k \) is defined as:

\[
PLI_{n,t} = \frac{1}{M} \sum_{k=0}^{M-1} \text{sgn} \left( \phi_n(k) - \phi_t(k) \right),
\]

\noindent where the \( \text{sgn} \) function assigns a value of -1, 0, or 1 based on whether the phase difference is negative, zero, or positive, respectively. The PLI ranges from 0 to 1, with values closer to 1 indicating significant phase lead or lag synchronization between the signals. Unlike PLV, PLI is insensitive to common sources and volume conduction, making it a robust measure for assessing true functional connectivity.

We meticulously trained and evaluated the EEGNet\cite{lawhern2018eegnet}, ShallowConvNet\cite{pan2016shallow}, and FBCNet\cite{mane2020multi} models for each dataset to ensure consistency and reliability in our results. The dataset was split into 70 \% for training, 20 \% for testing, and 10 \% for validation using a fixed random seed of 123. This stratified splitting method maintained a balanced distribution of classes across all partitions, preventing class imbalance issues.

For classification tasks, each model was trained for 100 epochs with a learning rate of \(1 \times 10^{-5}\). In contrast, for regression tasks, training was conducted for 50 epochs with a higher learning rate of \(1 \times 10^{-3}\). Across all experiments, a dropout rate of 0.5 and weight decay of \(5 \times 10^{-4}\) was employed to mitigate overfitting and enhance generalization performance. The hidden dimensions for both the convolutional layers and the multilayer perceptron layers in EEGNet, ShallowConvNet, and FBCNet were consistently set to 64. For loss functions, cross-entropy loss was utilized for classification tasks to effectively handle categorical outcomes, while mean absolute error was adopted for regression tasks to accurately measure prediction errors.

\section{RESULTS AND DISCUSSION}
\subsection{Connectivity Decoding Performance}

From the results in Table \ref{tab:freq_band_comparison}, it is evident that FBCNet generally outperforms the other models across most frequency bands, particularly in the \(\gamma\) band, where it achieves an accuracy of \(49.05\ \% \pm 1.45\ \%\). EEGNet also demonstrates competitive performance, especially in the \(\beta\) band, with an accuracy of \(48.52\ \% \pm 1.69\ \%\), which is comparable to FBCNet's performance. However, ShallowConvNet lags behind both EEGNet and FBCNet in most frequency bands, achieving its best performance in the \(\theta\) band (\(38.05\ \% \pm 2.21\ \%\)), which is still notably lower than the best results of EEGNet and FBCNet.

When comparing the models across frequency bands, several trends emerge. In the \(\theta\) band, EEGNet and FBCNet show strong performance, with FBCNet marginally outperforming EEGNet, while ShallowConvNet achieves its highest accuracy in this band but still falls short compared to the other models. Both EEGNet and FBCNet reach their highest accuracies in the \(\gamma\) band, with FBCNet slightly outperforming EEGNet, indicating that high-frequency information may be particularly valuable for these architectures.

Overall, the spatial convolutional neural network (CNN) module shows the most consistent and highest performance across frequency bands, particularly excelling in the \(\gamma\) band. EEGNet, while slightly trailing FBCNet in most cases, demonstrates robust performance, particularly in the \(\beta\) and \(\gamma\) bands. On the other hand, ShallowConvNet struggles to match the performance of EEGNet and FBCNet, particularly in the \(\delta\), \(\alpha\), and \(\beta\) bands. It may not be as well-suited for capturing frequency-specific information as the other models.

\subsection{Statistical Evaluation of Speech Paradigms}
Statistical analyses conducted under different conditions revealed significant differences (\(p < 0.05\), FDR-corrected), further supporting the results of the deep learning models. During the imagined speech, compared to the other three paradigms, connectivity values were low in the \(\theta\) band but high in the  \(\gamma\) band (\( t(9) = 2.45, \, p = 0.037 \)). The results for overt speech and whispered speech were statistically less significant, and perceived speech features, while relatively weak, were also statistically less significant. This suggests that the brain processes overt speech and whispered speech through similar neural circuits. In contrast, the distinct results observed in the imagined speech condition imply that imagined speech may activate different brain regions or processes than overt or whispered speech.

The \(\theta\) band is known to encompass signals related to dreaming and resting states. Studies have indicated that speech-related signals can also be present within \(\theta\), particularly during imagined speech\cite{yao2021reading}, which involves consciously imagining the act of speaking. The \(\theta\) band frequently manifests in the EEG channels F5, FT7, FT8, T8, TP10, F7, AF7, Fp1, Fp2, AF4, AF8, F4, and AF3, which are primarily located in the frontal and temporal lobes, with half of them situated in the anterior frontal lobe region, which is involved in higher-order cognitive processing \cite{jo2023neurophysiological}, during imagined speech. This observation is consistent with the activation of regions such as Broca's area, which is heavily involved in speech production. The significant features of the \(\theta\) band in these areas during imagined speech suggest that this frequency band plays a role in the neural processes underlying speech imagination.

High frequency, on the other hand, may include noise artifacts from physical activities such as eyebrow and head movements. Therefore, we conducted preprocessing aimed at minimizing noise before proceeding with the analysis. The \(\gamma\) band is typically associated with high-level cognitive tasks, suggesting that imagined speech might impose a lower cognitive load compared to actual speech. This could explain the lower \(\gamma\) PLV observed in imagined speech compared to overt speech (\( t(9) = 1.69, \, p = 0.048 \)).

\section{CONCLUSION}

This study examined the performance of deep learning models for EEG-based neural decoding, focusing on differentiating between speech paradigms (perceived, mimed, imagined, and overt speech) across various frequency bands. Spatial CNN module-based model consistently outperformed other models, particularly in the \(\gamma\) band, while EEGNet demonstrated strong results in the \(\beta\) and \(\gamma\) bands. ShallowConvNet, however, underperformed compared to EEGNet and FBCNet, indicating its limitations in frequency-specific decoding tasks. Statistical analyses revealed significant differences during the imagined speech, particularly in \(\theta\) and \(\gamma\) band connectivity, suggesting that imagined speech may activate distinct neural processes compared to overt or whispered speech, which was processed through similar circuits. These findings emphasize the importance of connectivity metrics, such as PLV and PLI, in decoding neural dynamics across different speech states. Overall, this study highlights the value of deep learning models and connectivity metrics in advancing EEG-based speech decoding, with potential applications in improving BCI communication systems. Future work should focus on further refining these models or developing graph-based models that can better learn topological features for real-world BCI applications.



\bibliography{Reference}
\bibliographystyle{IEEEtran}

\end{document}